\documentclass{article}

\usepackage[preprint]{neurips_2023}

\usepackage[utf8]{inputenc} 
\usepackage[T1]{fontenc}    
\usepackage{hyperref}       
\usepackage{url}            
\usepackage{booktabs}       
\usepackage{amsfonts}       
\usepackage{nicefrac}       
\usepackage{microtype}      
\usepackage{xcolor}         
\usepackage{graphicx}
\usepackage{framed}

\usepackage{caption}
\usepackage{subcaption}
\usepackage{wrapfig}
\usepackage{multirow}
\usepackage{amsmath}
\usepackage{amsfonts}
\usepackage{bm}
\usepackage{amssymb}
\usepackage{amsthm}
\usepackage{verbatim}
\usepackage{longtable}
\usepackage{arydshln}
\usepackage{algorithm}
\usepackage{algpseudocode}
\usepackage{mathtools, nccmath}

\newcommand{\ours}{\texttt{LLM-QAT}}
\newcommand*\samethanks[1][\value{footnote}]{\footnotemark[#1]}
\DeclarePairedDelimiter{\nint}\lfloor\rceil

\title{LLM-QAT: Data-Free Quantization Aware Training for Large Language Models}

\author{
  Zechun Liu\thanks{Equal contribution} \\
  Reality Labs, Meta Inc. \\
  \texttt{zechunliu@meta.com} \\
  \And
  Barlas Oğuz\samethanks \\
  Meta AI \\
  \texttt{barlaso@meta.com} \\
  \And
  Changsheng Zhao\\
  Reality Labs, Meta Inc. \\
  \texttt{cszhao@meta.com} \\
  \AND
  Ernie Chang \\
  Reality Labs, Meta Inc.\\
  \And
  Pierre Stock \\
  Meta AI \\
  \And
  Yashar Mehdad \\
  Meta AI \\
  \And
  Yangyang Shi\\
  Reality Labs, Meta Inc.\\
  \AND
  Raghuraman Krishnamoorthi \\
  Reality Labs, Meta Inc. \\
  \And
  Vikas Chandra \\
  Reality Labs, Meta Inc.\\
}
\begin{document}
\maketitle
\begin{abstract}
Several post-training quantization methods have been applied to large language models (LLMs), and have been shown to perform well down to 8-bits.  We find that these methods break down at lower bit precision, and investigate quantization aware training for LLMs (LLM-QAT) to push quantization levels even further.  We propose a data-free distillation method that leverages generations produced by the pre-trained model, which better preserves the original output distribution and allows quantizing any generative model independent of its training data, similar to post-training quantization methods.  In addition to quantizing weights and activations, we also quantize the KV cache, which is critical for increasing throughput and support long sequence dependencies at current model sizes.  We experiment with LLaMA models of sizes 7B, 13B, and 30B, at quantization levels down to 4-bits.  We observe large improvements over training-free methods, especially in the low-bit settings.
\end{abstract}

\section{Introduction}

Following GPT-3~\citep{gpt-3}, several families of large language models (LLMs) such as OPT~\citep{opt}, PALM~\citep{palm}, BLOOM~\citep{bloom}, Chinchilla~\citep{chinchilla} and LLaMA~\citep{llama} have established that increasing model size leads to improved model capabilities.  As a result, language models with tens of billions or even hundreds of billions of parameters have become the norm in today's AI landscape.  Despite the growing excitement around LLMs, serving such models to the benefit of billions of users faces significant hurdles due to their large computational cost and environmental footprint.

Fortunately, there has been an increasing effort to accurately quantize LLMs, with multiple recent works~\citep{xiao2022smoothquant,yao2022zeroquant} focusing on 8-bit post-training quantization of weights and activations and achieving little to no loss of accuracy. However, a 65 billion parameter LLaMA model still takes up 65GB of GPU memory with only its weights.  Moreover, the key-value (KV) cache holding activations for the attention layers can easily go into the tens of GBs, and is the throughput bottleneck in the long sequence length regime common in today's applications.  The aforementioned works do not consider KV cache quantization along with weight and activation quantization.  Unfortunately, SoTA post-training quantization methods dramatically degrade in quality when pushed beyond 8-bits.  For higher quantization levels, we find it necessary to resort to quantization-aware training (QAT).

To our knowledge, QAT for LLMs has not been investigated before.  This is understandable for two reasons.  First, LLM training is technically difficult and resource intensive.  Second, QAT needs training data, which for LLMs is difficult to obtain.  The sheer scale and diversity of pre-training data is itself an obstacle.  Pre-processing might be prohibitive, or worse, some data might simply not be available due to legal restrictions.  It is also incresingly common to train LLMs in multiple stages, involving instruction tuning and reinforcement learning~\citep{instruct-gpt}, which would be very difficult to replicate during QAT.  In this work, we side-step this issue by using generated data from the LLM itself for knowledge distillation.  This simple workaround, which we refer to as \emph{data-free} knowledge-distillation is applicable to any generative model independent of whether or not the original training data is available.  We show that this method is better able to preserve the original model's output distribution, even compared to training on large subsets of the original training set.  Moreover, we can successfully distill quantized models using only a small set (100k) of sampled data, thus keeping computational costs reasonable.  All of our experiments are conducted using a single 8-gpu training node.

As a result, we are able to distill the 7B, 13B and 30B LLaMA models with weights and KV cache quantized down to 4-bits. In this regard, our approach exhibits significant enhancements in quality compared to post-training quantization. Notably, larger models employing QAT outperform smaller models utilizing floating-point 16-bit representations, despite having similar model sizes. Furthermore, we have successfully quantized activations to 6-bit precision, surpassing what was possible with existing methods. For a comprehensive analysis of our experimental results and detailed ablations, please refer to Section \ref{sec:experiments}.

In summary, we present the first application of QAT to LLMs, resulting in the first accurate 4-bit quantized LLMs.  We also demonstrate quantizing the KV cache simultaneously with weights and activations, which is critical to alleviate throughput bottlenecks for long sequence generation.  All of this is achieved by a novel \emph{data-free} distillation method, which makes QAT practical for large pre-trained generative models.

\begin{figure}[t!]
    \centering
    \includegraphics[width=\linewidth]{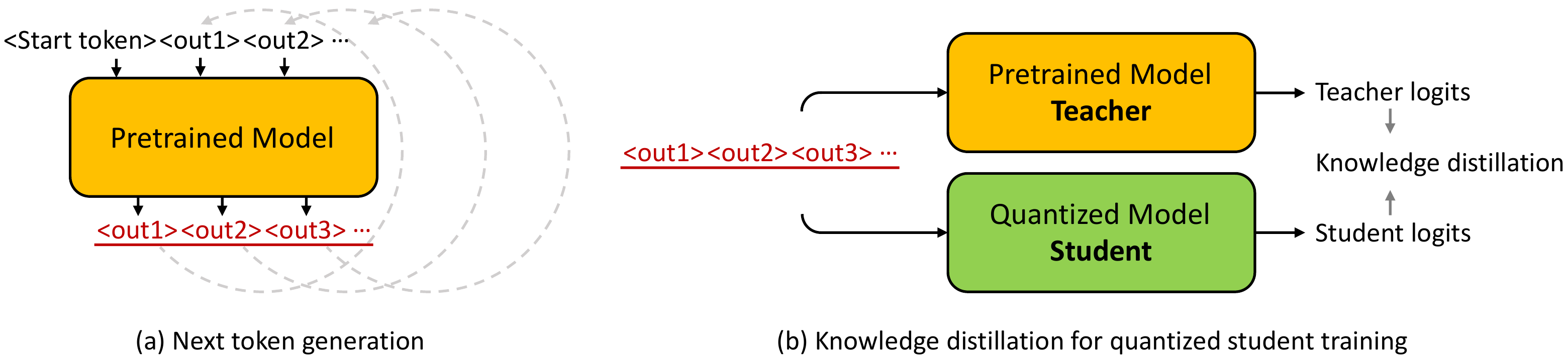}
    \caption{Overview of \ours{}. We generate data from the pretrained model with next token generation, which is sampled from top-k candidates. Then we use the generated data as input and the teacher model prediction as label to guide quantized model finetuning.}
    \label{fig:overview}
\end{figure}

\section{Method}
Quantizing large language models (LLMs) using quantization-aware training (QAT) is a nontrivial task with challenges in two key aspects. First, LLMs are pre-trained to excel in zero-shot generalization, and it is crucial to preserve this capability after quantization. Therefore, selecting an appropriate fine-tuning dataset is important.  If the QAT data is too narrow in domain or significantly different than the original pre-training distribution, this is likely to hurt model performance.  On the other hand, it is difficult to replicate the original training setup exactly, due to the scale and complexity of LLM training.  In Section~\ref{sec:generation}, we introduce \textit{data-free} quantization-aware training (QAT) which produces QAT data using next token data generation. This method demonstrates superior performance compared to using subsets of the original pre-training data. Second, LLMs exhibit unique weight and activation distributions characterized by a significant presence of outliers, which distinguishes them from smaller models. Consequently, the state-of-the-art quantization clipping methods for small models do not work out of the box for LLMs. In Section~\ref{sec:quantization}, we identify suitable quantizers for LLMs.

\subsection{Data-free Distillation}
\label{sec:generation}
In order to closely synthesize the distribution of the pre-training data with a limited amount of fine-tuning data, we proposed next token data generation from the original pre-trained model. As shown in Figure~\ref{fig:overview} (a), we randomize the first token $<\!\!start\!\!>$ from vocabulary and let the pre-trained model to generate the next token $<\!\!out1\!\!>$, then the generated token is appended to the start token for generating new output $<\!\!out2\!\!>$. We repeat this iterative procedure until we reach either the end of sentence token or the maximum generation length.

We test three different sampling strategies in the next token generation. The most straightforward way is to pick the top-1 candidate as the next token. However, the generated sentence lacks of diversity and will cyclically repeat several tokens. To address this issue, we instead stochastically sample the next token from the distribution using the SoftMax output of the pre-trained model as the probability. This sampling strategy yields more diverse sentences and greatly enhances the accuracy of the fine-tuned student model. Furthermore, we discover that the initial few tokens play a crucial role in determining the prediction trend. Therefore, it is important for them to have higher confidence.
In our generative process, we employ a hybrid sampling strategy that deterministically selects the top-1 predictions for the first 3\textasciitilde5 tokens and stochastically samples the remaining tokens. A detailed ablation study comparing different generated data and real data is presented in Section\ref{sec:ablation_data}.

\begin{figure}[t!]
    \centering
    \includegraphics[width=0.75\linewidth]{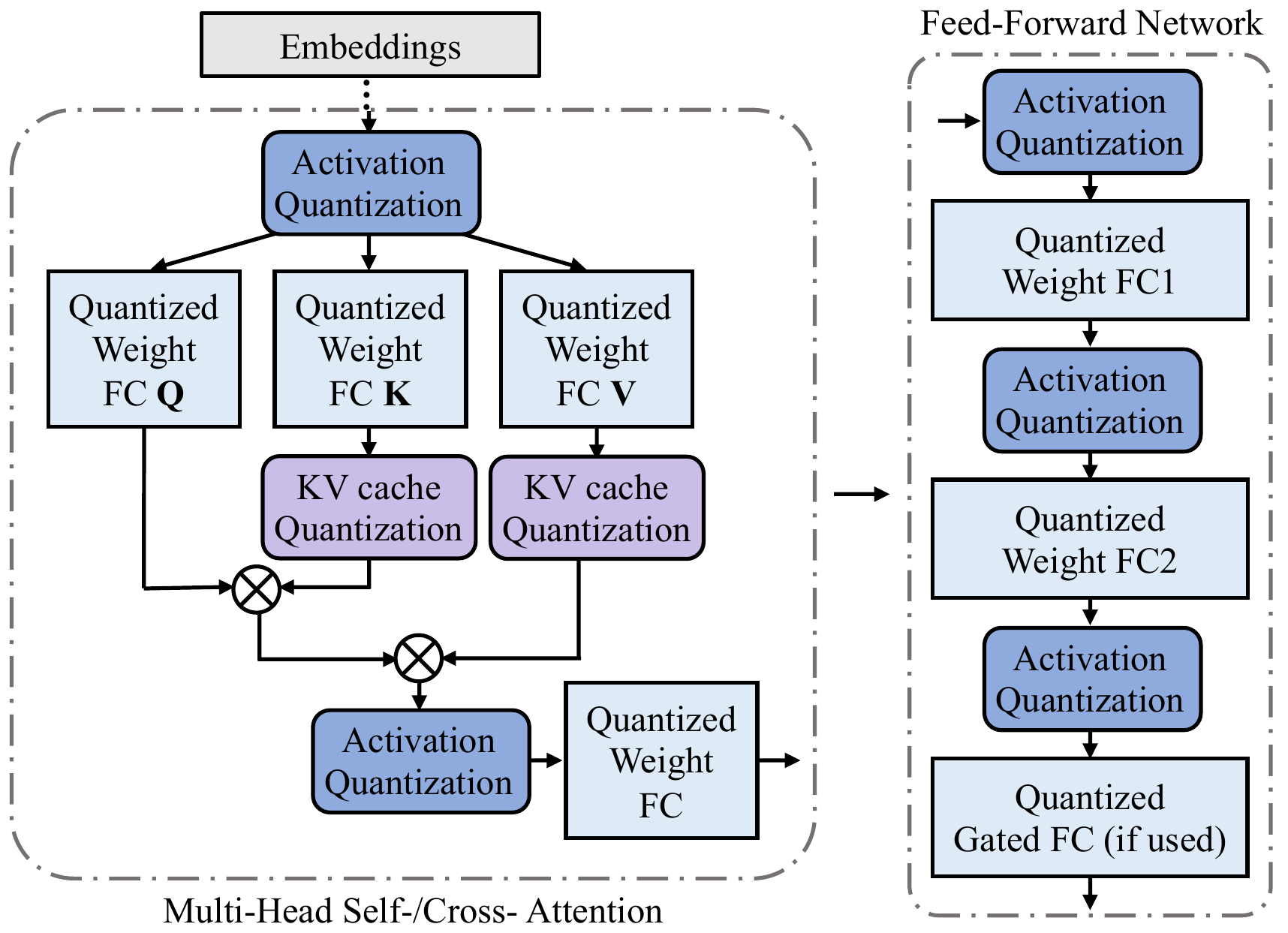}
    \caption{Overview of the quantized transformer in \ours{}. We quantize all the weights and input activations in fully-connected linear layers. The KV cache is also quantized if specified.}
    \label{fig:quantization}
\end{figure}
\subsection{Quantization-Aware Training}
\label{sec:quantization}
\subsubsection{Preliminaries}
In this work, we study linear quantization \textit{i.e.}, uniform quantization. Linear quantization can be categorized into two categories based on whether the real values are clipped or not: MinMax quantization, which preserves all value ranges, and clipping-based quantization.

In MinMax quantization, the quantization process can be formulated as:
\begin{equation}
\label{eq:minmax}
    \mathbf{X}_\mathbf{Q}^i = \alpha \mathbf{\hat{X}_Q}^i = \alpha \nint{\frac{\mathbf{X}_\mathbf{R}^i - \beta}{\alpha}} + \beta.
\end{equation}
Here $\mathbf{X}_\mathbf{Q}$ and $\mathbf{X}_\mathbf{R}$ denote the quantized and full-precision variables, respectively. $i$ refers to the $i$-th element in the tensor. $\alpha$ is the scaling factor and $\beta$ is the zero-point value.
For symmetric quantization, $\alpha = \frac{\max(|\mathbf{X}_\mathbf{R}|)}{2^{N-1} -1} , \ \beta = 0.$ And for asymmetric quantization, $\alpha = \frac{\max(\mathbf{X}_\mathbf{R}) - \min(\mathbf{X}_\mathbf{R})}{2^N -1} , \ \beta = \min(\mathbf{X}_\mathbf{R}).$

Compared to the MinMax Quantization, clipping the outliers can help improve the precision and allocate more bits to the intermediate values. Thus, many recent work~\citep{Q-BERT, TernaryBERT} adopts clipping-based quantization for transformer-based language models. The quantization can be formulated as:
\begin{equation}
    \mathbf{X}_\mathbf{Q}^i = \alpha \mathbf{\hat{X}_Q}^i = \alpha \nint{{\rm Clip}(\frac{\mathbf{X}_\mathbf{R}^i - \beta}{\alpha}, 0, 1)} + \beta.
\end{equation}
where the scale $\alpha$ and zero-point value $\beta$ can be calculated statistically or learned through gradients.

\subsubsection{Quantization for Large Language Models}

\begin{figure}[t!]
    \centering
    \includegraphics[width=0.85\linewidth]{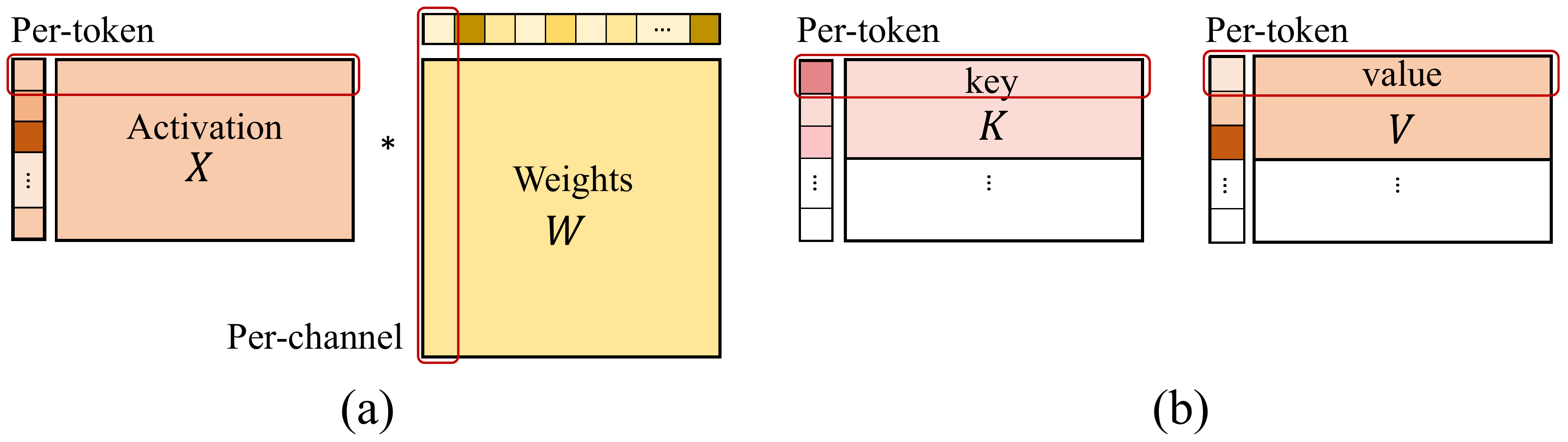}
    \caption{Illustration of (a) per-channel weight quantization and per-token activation quantization (b) per-token quantization for KV cache. The KV cache is updated by appending the current key and value to it. Thus we adopt per-token quantization for both key and value. }
    \label{fig:KV_quant}
\end{figure}

\noindent\textbf{Quantization function}
We illustrate our quantized transformer model in Figure~\ref{fig:quantization}. In line with the findings in~\citep{dettmers2022llm,xiao2022smoothquant}, we have also observed a significant presence of outliers in both the weights and activations of large language models (LLMs). These outliers have a notable impact on the quantization process, as they contribute to an increase in the quantization step size while diminishing the precision of intermediate values. Nevertheless, clipping these outliers during quantization proves detrimental to LLM performance. During the initial stages of training, any clipping-based method will lead to exceptionally high perplexity scores (\textit{i.e.,} $>10000$), causing a substantial loss of information that proves to be difficult to recover through fine-tuning. Therefore, we choose to retain these outliers instead. Moreover, we find that in the model with the gated linear unit (GLU), the activations are weights are mostly symmetrically distributed. Based on our analysis and empirical observations, we choose symmetric MinMax quantization for both weights and activations:
\begin{equation}
\label{eq:quantization}
    \mathbf{X}_\mathbf{Q}^i = \alpha \nint{\frac{\mathbf{X}_\mathbf{R}^i}{\alpha}}, \ \ \ \alpha = \frac{\max(|\mathbf{X}_\mathbf{R}|)}{2^{N-1} -1}
\end{equation}
Here $\mathbf{X}_\mathbf{Q}$ denotes the quantized weights or activations and $\mathbf{X}_\mathbf{R}$ denotes the real-valued weights or activations. To ensure efficient quantization, we adopt the per-token activation quantization and per-channel weight quantization as illustrated in Figure~\ref{fig:KV_quant} (a). For a comprehensive evaluation of the different quantizer choices, we provide the ablation study in Section~\ref{sec:ablation_quant}.

\textbf{Quantization-aware training for key-value cache}
In addition to weight and activation quantization, the key-value cache (KV cache) in large language models (LLMs) also consumes a non-negligible amount of memory. However, only a few previous works have addressed the KV cache quantization in LLMs, with the methods primarily limited to post-training quantization\citep{sheng2023high}.
In our study, we demonstrate that a similar quantization-aware training approach used for activation quantization can be employed to quantize the KV cache. As illustrated in Figure~\ref{fig:KV_quant}, we adopt per-token quantization in Eq.~\ref{eq:quantization}, given that the key and value are generated token by token. During the generation process, the current key and value are quantized, and their corresponding scaling factor is stored. During the training process for QAT, we apply quantization to the entire activation tensors of both the keys and values, as shown in Figure~\ref{fig:quantization}. By integrating the quantization function into the gradient calculation, we ensure effective training using quantized key-value pairs.

\textbf{Knowledge distillation}
We use cross-entropy based logits distillation for training the quantized student network from the full-precision pre-trained teacher network:
\begin{align}
    \mathcal{L}_{CE} = -\frac{1}{n}\sum_c\sum^n_{i=1} p_c^{\mathcal{T}}(X_i)\log(p_c^{\mathcal{S}}(X_i)),
\end{align}
Here $i$ denotes the $i^{th}$ sample in the current batch with $n$ total sentences. $c$ denotes the number of classes and in our case, it equals the size of the vocabulary. $\mathcal{T}$ and $\mathcal{S}$ are the teacher network and student network, respectively.

As discussed in Section~\ref{sec:generation}, in the data generation process, it is important to sample the next token from distribution rather than always selecting the top-1 candidate. By doing so, the next token does not necessarily represent the optimal label for training the student model, as the sampling introduces inherent noise. Consequently, we propose to utilize the predictions from the pre-trained model as soft labels, which provides more informative targets for guiding the training of the student model. We present a comprehensive ablation study in Section~\ref{sec:ablation_kd} to delve into the specifics of this approach.

\section{Experiments}\label{sec:experiments}
\begin{table}[btp]
\renewcommand\arraystretch{0.6}
\centering
\caption{Zero-shot performance on Common Sense Reasoning tasks.}
\label{tab:main}
\setlength{\tabcolsep}{1.2mm}
{\resizebox{0.95\textwidth}{!}{
\begin{tabular}{llccccccccccc}
\noalign{\vspace{0.1em}}\hline\noalign{\vspace{0.1em}}
& & & & \textbf{BoolQ} & \textbf{PIQA} & \textbf{SIQA} & \textbf{HellaSwag} & \textbf{WinoGrande} & \textbf{ARC-e} & \textbf{ARC-c} & \textbf{OBQA} & \textbf{Avg.} \\ 
& \textbf{Method} & \textbf{\#Bits} & \begin{tabular}[c]{@{}c@{}}\textbf{Size}$_\text{ (GB)}$\end{tabular} & ($\uparrow$) & ($\uparrow$)& ($\uparrow$)& ($\uparrow$)& ($\uparrow$)& ($\uparrow$)& ($\uparrow$)& ($\uparrow$)& ($\uparrow$) \\
\noalign{\vspace{0.1em}}\hline\noalign{\vspace{0.1em}}
\tiny{1}  & LLaMA-7B & 16-16-16 & 12.6 & 76.8 & 79.3 & 48.6 & 76.1 & 70.0 & 73.0 & 48.0 & 57.6 & 66.2 \\
            \noalign{\vspace{0.1em}}\hdashline[0.8pt/1pt]\noalign{\vspace{0.1em}} 
\tiny{2}  & RTN & 4-8-4 & 3.5 & 51.9 & 56.3 & 40.5 & 35.7 & 49.9 & 39.3 & 25.3 & 30.8 & 41.2 \\
\tiny{3}  & SmoothQuant & 4-8-4 & 3.5 & 54.7 & 55.4 & 41.1 & 38.9 & 51.5 & 43.9 & 27.7 & 32.0 & 43.2 \\
\tiny{4}  & \ours{} & 4-8-4 & 3.5 & 69.5 & 75.4 & 46.6 & 69.2 & 64.6 & 66.0 & 43.8 & 50.6 & \textbf{60.7 }\\
            \noalign{\vspace{0.1em}}\hdashline[0.8pt/1pt]\noalign{\vspace{0.1em}} 
\tiny{5}  & RTN & 4-8-8 & 3.5 & 67.8 & 76.6 & 47.2 & 71.4 & 67.2 & 67.4 & 45.6 & 51.2 & 61.8 \\
\tiny{6}  & SmoothQuant & 4-8-8 & 3.5 & 71.0 & 76.0 & 45.4 & 67.8 & 66.0 & 67.4 & 42.8 & 47.0 & 60.4 \\
\tiny{7}  & \ours{} & 4-8-8 & 3.5 & 74.6 & 77.5 & 48.3 & 73.5 & 67.7 & 70.2 & 45.6 & 56.2 & \textbf{64.2} \\
            \noalign{\vspace{0.1em}}\hdashline[0.8pt/1pt]\noalign{\vspace{0.1em}} 
\tiny{8 } & RTN & 4-6-16 & 3.5 & 62.4 & 74.5 & 46.8 & 67.9 & 64.5 & 64.6 & 41.5 & 49.0 & 58.9 \\
\tiny{9 } & SmoothQuant & 4-6-16 & 3.5 & 68.8 & 73.9 & 44.5 & 65.7 & 65.3 & 66.0 & 43.6 & 48.0 & 59.5 \\
\tiny{10} & \ours{} & 4-6-16 & 3.5 & 72.9 & 76.8 & 47.9 & 72.4 & 68.3 & 68.8 & 44.2 & 53.2 & \textbf{63.1} \\
            \noalign{\vspace{0.1em}}\hdashline[0.8pt/1pt]\noalign{\vspace{0.1em}} 
\tiny{11} & RTN & 4-8-16 & 3.5 & 67.6 & 77.4 & 47.1 & 71.6 & 66.9 & 67.1 & 45.8 & 52.0 & 61.9 \\
\tiny{12} & SmoothQuant & 4-8-16 & 3.5 & 70.2 & 76.4 & 44.8 & 68.1 & 66.0 & 67.3 & 42.9 & 49.0 & 60.6 \\
\tiny{13} & \ours{} & 4-8-16 & 3.5 & 74.8 & 77.8 & 48.6 & 73.6 & 69.0 & 69.7 & 45.8 & 55.8 & \textbf{64.4 }\\
            \noalign{\vspace{0.1em}}\hdashline[0.8pt/1pt]\noalign{\vspace{0.1em}} 
\tiny{14} & RTN & 4-16-16 & 3.5 & 71.2 & 77.3 & 47.6 & 72.7 & 66.9 & 68.8 & 46.4 & 52.8 & 63.0 \\
\tiny{15} & GPTQ & 4-16-16 & 3.5 & 67.7 & 76.0 & 46.8 & 69.4 & 66.7 & 66.9 & 43.0 & 50.6 & 60.9 \\
\tiny{16} & \ours{} & 4-16-16 & 3.5 & 75.5 & 78.3 & 48.4 & 74.0 & 69.0 & 70.0 & 45.0 & 55.4 & \textbf{64.4} \\
            \noalign{\vspace{0.1em}}\hdashline[0.8pt/1pt]\noalign{\vspace{0.1em}} 
\tiny{17} & RTN & 8-8-4 & 6.5 & 54.7 & 59.4 & 43.1 & 45.6 & 57.4 & 51.2 & 29.6 & 37.8 & 47.4 \\
\tiny{18} & SmoothQuant & 8-8-4 & 6.5 & 60.7 & 67.5 & 44.9 & 58.3 & 58.6 & 57.5 & 36.9 & 43.6 & 53.5 \\
\tiny{19} & \ours{} & 8-8-4 & 6.5 & 71.1 & 75.6 & 47.3 & 71.8 & 66.3 & 67.1 & 43.6 & 50.0 & \textbf{61.6} \\
            \noalign{\vspace{0.1em}}\hdashline[0.8pt/1pt]\noalign{\vspace{0.1em}} 
\tiny{20} & RTN & 8-8-8 & 6.5 & 76.4 & 79.5 & 48.7 & 75.5 & 69.5 & 72.3 & 46.6 & 56.0 & 65.6 \\
\tiny{21} & SmoothQuant & 8-8-8 & 6.5 & 76.1 & 79.6 & 48.7 & 76.2 & 70.1 & 73.7 & 48.7 & 57.0 & \textbf{66.3} \\
\tiny{22} & \ours{} & 8-8-8 & 6.5 & 76.0 & 79.6 & 48.5 & 75.7 & 69.4 & 73.1 & 48.2 & 57.4 & 66.0 \\
            \noalign{\vspace{0.1em}}\hdashline[0.8pt/1pt]\noalign{\vspace{0.1em}} 
\tiny{23} & RTN & 8-8-16 & 6.5 & 76.4 & 79.1 & 48.3 & 75.7 & 70.5 & 72.8 & 46.5 & 55.6 & 65.6 \\
\tiny{24} & SmoothQuant & 8-8-16 & 6.5 & 76.2 & 79.5 & 48.6 & 76.1 & 70.5 & 73.2 & 47.7 & 57.2 & \textbf{66.1} \\
\tiny{25} & \ours{} & 8-8-16 & 6.5 & 76.3 & 79.4 & 48.7 & 75.6 & 69.7 & 72.3 & 47.6 & 56.2 & 65.7 \\

            \noalign{\vspace{0.1em}}\hline\noalign{\vspace{0.1em}}
\tiny{26} & LLaMA-13B & 16-16-16 & 24.2 & 78.1 & 80.0 & 50.5 & 79.2 & 73.6 & 74.5 & 52.6 & 55.0 & 68.0 \\
            \noalign{\vspace{0.1em}}\hdashline[0.8pt/1pt]\noalign{\vspace{0.1em}} 
\tiny{27} & RTN & 4-8-4 & 6.5 & 54.0 & 59.2 & 41.9 & 41.6 & 55.9 & 45.0 & 27.0 & 33.2 & 44.7 \\
\tiny{28} & SmoothQuant & 4-8-4 & 6.5 & 63.0 & 65.3 & 42.2 & 50.6 & 54.1 & 49.6 & 30.3 & 34.2 & 48.7 \\
\tiny{29} & \ours{} & 4-8-4 & 6.5 & 72.0 & 76.8 & 49.2 & 73.6 & 66.5 & 69.3 & 46.9 & 52.8 & \textbf{63.4} \\
            \noalign{\vspace{0.1em}}\hdashline[0.8pt/1pt]\noalign{\vspace{0.1em}} 
\tiny{30} & RTN & 4-8-8 & 6.5 & 76.2 & 78.8 & 49.3 & 76.2 & 69.9 & 72.2 & 50.7 & 56.8 & 66.3 \\
\tiny{31} & SmoothQuant & 4-8-8 & 6.5 & 72.5 & 77.1 & 47.2 & 74.3 & 69.5 & 67.4 & 43.3 & 53.4 & 63.1 \\
\tiny{32} & \ours{} & 4-8-8 & 6.5 & 77.5 & 79.1 & 48.6 & 77.5 & 70.6 & 73.0 & 51.9 & 56.2 & \textbf{66.8} \\
            \noalign{\vspace{0.1em}}\hdashline[0.8pt/1pt]\noalign{\vspace{0.1em}} 
\tiny{33} & RTN & 4-6-16 & 6.5 & 71.8 & 74.1 & 47.7 & 70.2 & 65.1 & 69.3 & 44.1 & 45.6 & 61.0 \\
\tiny{34} & SmoothQuant & 4-6-16 & 6.5 & 70.6 & 76.3 & 47.9 & 73.1 & 68.5 & 65.9 & 43.3 & 52.6 & 62.3 \\
\tiny{35} & \ours{} & 4-6-16 & 6.5 & 75.4 & 79.3 & 48.4 & 76.5 & 69.2 & 73.1 & 48.6 & 53.4 & \textbf{65.5} \\
            \noalign{\vspace{0.1em}}\hdashline[0.8pt/1pt]\noalign{\vspace{0.1em}} 
\tiny{36} & RTN & 4-8-16 & 6.5 & 76.8 & 79.1 & 49.1 & 76.3 & 70.5 & 72.6 & 49.8 & 56.6 & 66.4 \\
\tiny{37} & SmoothQuant & 4-8-16 & 6.5 & 72.5 & 77.9 & 47.6 & 74.2 & 69.7 & 68.2 & 45.0 & 54.2 & 63.7 \\
\tiny{38} & \ours{} & 4-8-16 & 6.5 & 77.7 & 79.3 & 48.4 & 77.5 & 70.6 & 73.5 & 53.0 & 57.4 & \textbf{67.2} \\
            \noalign{\vspace{0.1em}}\hdashline[0.8pt/1pt]\noalign{\vspace{0.1em}} 
\tiny{39} & RTN & 4-16-16 & 6.5 & 77.4 & 79.1 & 49.2 & 76.8 & 70.5 & 72.6 & 51.2 & 54.2 & 66.4 \\
\tiny{40} & GPTQ & 4-16-16 & 6.5 & 78.0 & 79.8 & 49.2 & 77.7 & 72.6 & 73.2 & 50.6 & 55.4 & \textbf{67.1} \\
\tiny{41} & \ours{} & 4-16-16 & 6.5 & 77.7 & 79.4 & 49.1 & 77.7 & 71.5 & 72.8 & 52.0 & 53.8 & 66.7 \\
            \noalign{\vspace{0.1em}}\hdashline[0.8pt/1pt]\noalign{\vspace{0.1em}} 
\tiny{42} & RTN & 8-8-4 & 12.4 & 65.8 & 66.2 & 43.9 & 56.7 & 57.3 & 58.2 & 34.5 & 42.6 & 53.2 \\
\tiny{43} & SmoothQuant & 8-8-4 & 12.4 & 66.6 & 71.7 & 44.8 & 61.1 & 61.0 & 63.4 & 38.3 & 43.6 & 56.3 \\
\tiny{44} & \ours{} & 8-8-4 & 12.4 & 74.9 & 78.3 & 48.0 & 75.7 & 68.9 & 71.9 & 51.1 & 54.2 & \textbf{65.4} \\
            \noalign{\vspace{0.1em}}\hdashline[0.8pt/1pt]\noalign{\vspace{0.1em}} 
\tiny{45} & RTN & 8-8-8 & 12.4 & 77.8 & 80.0 & 50.8 & 78.9 & 72.6 & 74.5 & 52.1 & 55.6 & 67.8 \\
\tiny{46} & SmoothQuant & 8-8-8 & 12.4 & 78.3 & 80.3 & 50.8 & 79.2 & 73.2 & 74.8 & 52.4 & 55.4 & \textbf{68.0} \\
\tiny{47} & \ours{} & 8-8-8 & 12.4 & 78.7 & 80.4 & 50.1 & 79.1 & 73.2 & 74.8 & 51.7 & 55.4 & 67.9 \\
            \noalign{\vspace{0.1em}}\hdashline[0.8pt/1pt]\noalign{\vspace{0.1em}} 
\tiny{48} & RTN & 8-8-16 & 12.4 & 77.8 & 80.1 & 50.6 & 78.9 & 73.5 & 74.9 & 51.9 & 56.4 & \textbf{68.0} \\
\tiny{49} & SmoothQuant & 8-8-16 & 12.4 & 78.7 & 80.0 & 50.6 & 79.1 & 73.4 & 74.8 & 51.4 & 56.0 & \textbf{68.0} \\
\tiny{50} & \ours{} & 8-8-16 & 12.4 & 78.5 & 80.4 & 50.6 & 79.0 & 72.8 & 74.2 & 52.9 & 55.8 & \textbf{68.0} \\

            \noalign{\vspace{0.1em}}\hline\noalign{\vspace{0.1em}}
\tiny{51} &  LLaMA-30B & 16-16-16 & 60.6 & 83.2 & 82.1 & 50.4 & 82.9 & 75.6 & 80 & 58 & 59.3 & 71.4 \\
 \noalign{\vspace{0.1em}}\hdashline[0.8pt/1pt]\noalign{\vspace{0.1em}} 
\tiny{52} & RTN & 4-8-4 & 15.7 & 56.9 & 56.2 & 40.2 & 39.6 & 50.0 & 40.6 & 26.4 & 29.8 & 42.5 \\
\tiny{53} & SmoothQuant & 4-8-4 & 15.7 & 56.6 & 55.0 & 39.9 & 33.8 & 49.9 & 38.8 & 24.5 & 27.2 & 40.7 \\
\tiny{54} & \ours{} & 4-8-4 & 15.7 & 80.5 & 80.3 & 49.7 & 80.2 & 75.2 & 78.2 & 56.0 & 59.2 & \textbf{69.9} \\
            \noalign{\vspace{0.1em}}\hdashline[0.8pt/1pt]\noalign{\vspace{0.1em}} 
\tiny{55} & RTN & 4-8-8 & 15.7 & 78.8 & 79.9 & 49.0 & 80.2 & 75.2 & 78.4 & 54.4 & 57.2 & 69.1 \\
\tiny{56} & SmoothQuant & 4-8-8 & 15.7 & 74.9 & 79.5 & 47.1 & 76.9 & 70.6 & 76.5 & 54.5 & 55.0 & 66.9 \\
\tiny{57} & \ours{} & 4-8-8 & 15.7 & 81.3 & 80.9 & 50.4 & 81.3 & 76.3 & 80.3 & 56.5 & 57.0 & \textbf{70.5} \\
            \noalign{\vspace{0.1em}}\hdashline[0.8pt/1pt]\noalign{\vspace{0.1em}} 
\tiny{58} & RTN & 4-6-16 & 15.7 & 64.5 & 57.0 & 42.1 & 48.9 & 55.4 & 39.3 & 27.0 & 32.2 & 45.8 \\
\tiny{59} & SmoothQuant & 4-6-16 & 15.7 & 75.0 & 77.6 & 46.6 & 73.8 & 69.1 & 74.5 & 52.9 & 50.6 & 65.0 \\
\tiny{60} & \ours{} & 4-6-16 & 15.7 & 78.8 & 80.3 & 50.3 & 79.9 & 75.1 & 77.0 & 54.4 & 59.0 & \textbf{69.4} \\
            \noalign{\vspace{0.1em}}\hdashline[0.8pt/1pt]\noalign{\vspace{0.1em}} 
\tiny{61} & RTN & 4-8-16 & 15.7 & 79.1 & 79.6 & 49.5 & 80.4 & 74.9 & 78.3 & 53.7 & 57.2 & 69.1 \\
\tiny{62} & SmoothQuant & 4-8-16 & 15.7 & 76.0 & 79.8 & 48.2 & 77.0 & 71.6 & 76.4 & 55.6 & 54.2 & 67.3 \\
\tiny{63} & \ours{} & 4-8-16 & 15.7 & 80.6 & 80.8 & 50.1 & 81.2 & 75.8 & 79.7 & 56.3 & 56.3 & \textbf{70.1} \\
            \noalign{\vspace{0.1em}}\hdashline[0.8pt/1pt]\noalign{\vspace{0.1em}} 
\tiny{64} & RTN & 4-16-16 & 15.7 & 80.8 & 80.1 & 49.8 & 81.6 & 75.8 & 79.3 & 55.8 & 57.2 & 70.1 \\
\tiny{65} & GPTQ & 4-16-16 & 15.7 & 81.0 & 81.6 & 49.7 & 82.2 & 74.3 & 79.6 & 56.1 & 58.2 & \textbf{70.3} \\
\tiny{66} & \ours{} & 4-16-16 & 15.7 & 81.8 & 81.0 & 49.7 & 81.8 & 75.1 & 79.4 & 56.8 & 54.9 & 70.1 \\
            \noalign{\vspace{0.1em}}\hdashline[0.8pt/1pt]\noalign{\vspace{0.1em}} 
\tiny{67} & RTN & 8-8-4 & 30.7 & 59.8 & 64.5 & 42.7 & 51.8 & 55.0 & 52.2 & 33.2 & 38.0 & 49.6 \\
\tiny{68} & SmoothQuant & 8-8-4 & 30.7 & 58.9 & 63.7 & 43.5 & 54.8 & 55.2 & 55.3 & 33.6 & 40.2 & 50.7 \\
\tiny{69} & \ours{} & 8-8-4 & 30.7 & 81.2 & 81.6 & 50.1 & 81.1 & 73.6 & 78.5 & 55.7 & 55.7 & \textbf{69.7} \\
            \noalign{\vspace{0.1em}}\hdashline[0.8pt/1pt]\noalign{\vspace{0.1em}} 
\tiny{70} & RTN & 8-8-8 & 30.7 & 82.2 & 81.2 & 49.4 & 81.9 & 75.6 & 79.6 & 57.4 & 58.2 & 70.7 \\
\tiny{71} & SmoothQuant & 8-8-8 & 30.7 & 82.5 & 82.3 & 50.2 & 82.8 & 75.9 & 80.3 & 56.9 & 57.8 & \textbf{71.1} \\
\tiny{72} & \ours{} & 8-8-8 & 30.7 & 82.2 & 81.3 & 51.0 & 82.3 & 75.0 & 80.2 & 57.0 & 57.2 & 70.8 \\
            \noalign{\vspace{0.1em}}\hdashline[0.8pt/1pt]\noalign{\vspace{0.1em}} 
\tiny{73} & RTN & 8-8-16 & 30.7 & 82.3 & 81.6 & 50.2 & 81.7 & 75.9 & 79.7 & 56.7 & 59.0 & 70.9 \\
\tiny{74} & SmoothQuant & 8-8-16 & 30.7 & 82.8 & 81.9 & 50.3 & 82.7 & 76.3 & 80.2 & 57.7 & 58.4 & \textbf{71.3} \\
\tiny{75} & \ours{} & 8-8-16 & 30.7 & 82.4 & 81.4 & 50.3 & 82.5 & 76.0 & 80.0 & 57.2 & 56.8 & 70.8 \\
\hline
\end{tabular}}}
\vspace{-2em}
\end{table}

We assess the effectiveness of our approach by conducting experiments on LLaMA-7B/13B/30B models and presenting results on various tasks. Specifically, we report the zero-shot performance on Common Sense Reasoning tasks such as BoolQ~\citep{clark2019boolq}, PIQA~\citep{bisk2020piqa}, SIQA~\citep{sap2019siqa}, HellaSwag~\citep{zellers2019hellaswag}, WinoGrande~\citep{sakaguchi2021winogrande}, ARC~\citep{clark2018arc}, and OBQA~\citep{mihaylov2018obqa}. We also assess the few-shot performance on TriviaQA~\citep{joshi2017triviaqa} and MMLU~\citep{hendrycks2020mmlu} datasets, along with perplexity scores on WikiText2~\citep{merity2016wiki2} and C4~\citep{raffel2020c4} datasets.

\subsection{Experimental Settings}
In our quantized network training process, we initialize the model with a pre-trained model and employ it as the teacher for knowledge distillation. To optimize the model, we utilize the AdamW~\citep{loshchilov2017decoupled} optimizer with zero weight decay. Each GPU is assigned a batch size of 1, and the learning rate is set to 2e-5, following a cosine learning-rate decay strategy. For data generation, we utilize the LLaMA-7B model, and the maximum length of generated sequences is set to 1024.

\subsection{Main Results}
We consider three post-training quantization (PTQ) methods, round-to-nearest (RTN), GPT-Q~\citep{frantar2022gptq} and SmoothQuant~\citep{xiao2022smoothquant} as baselines.  We compare to them in several different settings, where the weights, activations and KV cache values are quantized to different levels (denoted as W-A-KV).  Different PTQ methods perform well in different settings, and we compare our method to the best PTQ result in each setting.

Table~\ref{tab:main}, table~\ref{tab:ppl} and table 7 (in Appendix) give the comparisons of the proposed QAT methods with SOTA PTQ methods for LLMs on Zero-shot tasks on Common Sense Reasoning tasks, perplexity evaluation on Wiki2 and C4 and few shot exact match on the MMLU and TriviaQA benchmarks respectively.  The perplexity evaluations verify whether the quantize models are able to preserve the output distribution of the model on a diverse sample of its training domains.  The zero-shot and few-shot evaluations measure if the model's capabilities on downstream tasks are retained.

The trends in each table are similar.  All methods tend to do well in the 8-bit setting across all model sizes.  This holds even when the KV cache is also quantized to 8-bits, together with weights and activations.  However, when either of these three values are quantized to less than 8-bits, PTQ methods result in accuracy loss, whereas LLM-QAT holds up much better.  For example in the 8-8-4 setting, 30B LLM-QAT achieves an average zero-shot accuracy of 69.7, compared to 50.7 with SmoothQuant (Table~\ref{tab:main}, rows 68-69).  The difference is smaller in the 4-8-8 setting, however LLM-QAT still outperforms the best PTQ method (RTN in this case) by 1.4 points (rows 55, 57).  In the 4-8-4 setting, where both weights and the KV cache are quantized to 4 bits, all PTQ methods produce poor results, whereas LLM-QAT achieves 69.9, only trailing the full precision model by 1.5 points on average.  LLM-QAT also works reasonably well for 6-bit activation quantization.  While this setting might not be currently practical due to lack of hardware support, it's a promising data point for sub-8-bit computation for LLMs.  Unfortunately 4-bit activation quantization did not work well for the settings that we tried (see Section \ref{sec:smoothquant}).

One important question for practitioners is whether to use a small model at full precision, or a larger quantized model of similar inference cost.  While the exact trade-offs can vary based on several factors, we can make several recommendations based on our results.  First, 8-bit quantization should be preferred over smaller full precision models, and PTQ methods are sufficient for this case.  An 8-8-8 30B quantized model outperforms a 13B model of similar size, and should have lower latency and higher throughput in practice.  This also holds for an 8-bit 13B model compared with a 16-bit 7B model.  Furthermore, 4-bit models quantized using LLM-QAT should be preferred over 8-bit models of similar size.  For instance a 4-8-4 LLM-QAT 30B outperforms an 8-bit LLaMA-13B, and a 4-8-8 LLM-QAT 13B is better than an 8-bit LLaMA-7B.  As a result, we recommend 4-bit LLM-QAT models for the best efficiency-accuracy tradeoff.

\begin{table}[t]
\renewcommand\arraystretch{0.6}
\centering
\caption{Perplexity evaluation results on WikiText~\cite{merity2016wiki2} and C4~\cite{raffel2020c4}}
\label{tab:ppl}
\setlength{\tabcolsep}{1.5mm}
{\resizebox{0.98\textwidth}{!}{
\begin{tabular}{lc:lcc:lcc:lcc}
\vspace{-1.2em}
& & & & & & &\\
& & & & & & &\\
\noalign{\vspace{0.1em}}\hline\noalign{\vspace{0.1em}}
& & &\multicolumn{2}{c}{\textbf{Perplexity}} & & \multicolumn{2}{c}{\textbf{Perplexity}}  & & \multicolumn{2}{c}{\textbf{Perplexity}} \\
& \textbf{\#Bits} & \textbf{Method} & \textbf{C4} ($\downarrow$) & \textbf{Wiki2} ($\downarrow$) & \textbf{Method} & \textbf{C4}  ($\downarrow$) & \textbf{Wiki2} ($\downarrow$)  & \textbf{Method} & \textbf{C4}  ($\downarrow$) & \textbf{Wiki2} ($\downarrow$)  \\ 
\noalign{\vspace{0.1em}}\hline\noalign{\vspace{0.1em}}
\tiny{1} &  16-16-16 & LLaMA-7B & 7.2 & 10.4 & LLaMA-13B & 6.7 & 9.7 & LLaMA-30B & 6.0 & 7.0\\
          \noalign{\vspace{0.1em}}\hdashline[0.8pt/1pt]\noalign{\vspace{0.1em}} 
\tiny{2} &  4-8-4 & RTN & 55.1 & 151.4 & RTN & 25.0 & 103.6 & RTN &  8.2 & 8.9 \\
\tiny{3} &  4-8-4 & SmoothQuant & 81.1 & 163.6 & SmoothQuant & 26.0 & 60.1 & SmoothQuant & 10.6 & 12.0\\
\tiny{4} &  4-8-4 & \ours{} & \textbf{8.6} & \textbf{11.6} & \ours{} & \textbf{7.6} & \textbf{10.2} & \ours{} & \textbf{7.3} & \textbf{7.7} \\
          \noalign{\vspace{0.1em}}\hdashline[0.8pt/1pt]\noalign{\vspace{0.1em}} 
\tiny{5} &  4-8-8 & RTN & 8.4 & 13.9 & RTN & 7.3 & 12.5 & RTN & 7.4 & 8.2 \\
\tiny{6} &  4-8-8 & SmoothQuant & 9.1 & 13.7 & SmoothQuant & 8.8 & 12.5 & SmoothQuant & 8.7 & 9.8 \\
\tiny{7} &  4-8-8 & \ours{} & \textbf{7.5} & \textbf{11.2} & \ours{} & \textbf{6.8} & \textbf{10.0} & \ours{} & \textbf{6.9} & \textbf{7.5} \\
          \noalign{\vspace{0.1em}}\hdashline[0.8pt/1pt]\noalign{\vspace{0.1em}} 
\tiny{8 } & 4-6-16 & RTN & 10.5 & 20.0 & RTN & 11.3 & 32.7 & RTN & 11.4 & 15.4 \\
\tiny{9 } & 4-6-16 & SmoothQuant & 9.9 & 14.7 & SmoothQuant & 9.1 & 13.6 & SmoothQuant & 8.7 & 12.5 \\
\tiny{10} & 4-6-16 & \ours{} & \textbf{7.7} & \textbf{10.8} & \ours{} & \textbf{7.1} & \textbf{10.5} & \ours{} & \textbf{7.3} & \textbf{7.9} \\
          \noalign{\vspace{0.1em}}\hdashline[0.8pt/1pt]\noalign{\vspace{0.1em}} 
\tiny{11} & 4-8-16 & RTN & 8.6 & 14.0 & RTN & 7.5 & 12.5 & RTN & 7.4 & 8.2 \\
\tiny{12} & 4-8-16 & SmoothQuant & 9.1 & 13.7 & SmoothQuant & 8.7 & 12.6 & SmoothQuant & 8.7 & 9.8 \\
\tiny{13} & 4-8-16 & \ours{} & \textbf{7.4} & \textbf{10.9} & \ours{} & \textbf{6.8} & \textbf{10.0} & \ours{} & \textbf{6.9} & \textbf{7.5}  \\
                    \noalign{\vspace{0.1em}}\hdashline[0.8pt/1pt]\noalign{\vspace{0.1em}} 
\tiny{14} & 4-16-16 & RTN & 8.5 & 14.4 & RTN & 7.3 & 11.9 & RTN & 7.0 & 7.7 \\
\tiny{15} & 4-16-16 & GPTQ & 8.4 & 17.4 & GPTQ & 6.8 & 10.7 & GPTQ & \textbf{6.2} & 7.9 \\
\tiny{16} & 4-16-16 & \ours{} & \textbf{7.4} & \textbf{10.9} & \ours{} & \textbf{6.5} & \textbf{9.6} & \ours{} & 6.5 & \textbf{7.3} \\
          \noalign{\vspace{0.1em}}\hdashline[0.8pt/1pt]\noalign{\vspace{0.1em}} 
\tiny{17} & 8-8-4 & RTN & 42.1 & 105.1 & RTN & 15.4 & 43.4 & RTN & 7.0 & 7.8 \\
\tiny{18} & 8-8-4 & SmoothQuant & 30.8 & 77.9 & SmoothQuant & 13.9 & 40.9 & SmoothQuant & \textbf{6.7} & 7.5 \\
\tiny{19} & 8-8-4 & \ours{} & \textbf{7.6} & \textbf{10.2} & \ours{} & \textbf{7.5} & \textbf{11.3} & \ours{} & 6.8 & \textbf{7.4} \\
          \noalign{\vspace{0.1em}}\hdashline[0.8pt/1pt]\noalign{\vspace{0.1em}} 
\tiny{20} & 8-8-8 & RTN & 7.1 & 10.7 & RTN & 6.6 & 10.0 & RTN & 6.3 & 7.3 \\
\tiny{21} & 8-8-8 & SmoothQuant & \textbf{7.0} & 10.5 & SmoothQuant & \textbf{6.5} & 9.8 & SmoothQuant & \textbf{6.1} & \textbf{7.1} \\
\tiny{22} & 8-8-8 & \ours{} & \textbf{7.0} & \textbf{10.3} & \ours{} & 7.0 & \textbf{9.4} & \ours{} & 6.3 & \textbf{7.1} \\
          \noalign{\vspace{0.1em}}\hdashline[0.8pt/1pt]\noalign{\vspace{0.1em}} 
\tiny{23} & 8-8-16 & RTN & 7.3 & 10.7 & RTN & 6.8 & 10.1 & RTN & 6.3 & 7.3 \\
\tiny{24} & 8-8-16 & SmoothQuant & \textbf{7.0} & 10.5 & SmoothQuant & \textbf{6.5} & 9.7 & SmoothQuant & \textbf{6.1} & \textbf{7.1}\\
\tiny{25} & 8-8-16 & \ours{} & \textbf{7.0} & \textbf{10.3} & \ours{} & \textbf{6.5} & \textbf{9.5} & \ours{} & 6.3 & \textbf{7.1} \\
\hline
\end{tabular}}}
\end{table}
\subsection{Ablation}
We conduct the ablation study regarding the data choice, quantization methods, and knowledge distillation methods in Section~\ref{sec:ablation_data}, Section~\ref{sec:ablation_quant} and Section~\ref{sec:ablation_kd}, respectively. We report both the perplexity scores on WikiText2~\citep{merity2016wiki2}/C4~\citep{raffel2020c4} datasets and the performance on zero-shot common sense reasoning tasks.

\subsubsection{Data Choice}
\label{sec:ablation_data}
In Table~\ref{tab:ablation_data}, we observe that WikiText (Merity et al., 2016), which is constructed using text extracted from Wikipedia, does not encompass all the information utilized during pre-training. Consequently, a model fine-tuned solely on WikiText tends to overfit on this specific dataset and struggles to generalize well to other datasets. On the other hand, the Crawled Corpus (C4) dataset (Raffel et al., 2020) comprises hundreds of gigabytes of clean English text collected from the web. Fine-tuning the model on C4 yields reasonable transfer accuracy when evaluated on the WikiText dataset. However, it exhibits poor accuracy when tasked with zero-shot inference tasks.

Compared to the existing data, the model fine-tuned on generated data demonstrates superior generalizability, particularly in zero-shot tasks. Moreover, the data generated through sampling from the distribution exhibits greater diversity compared to the data generated without sampling. This enhanced diversity leads to significantly improved performance across all tasks.
\begin{table}[t]
\renewcommand\arraystretch{0.6}
\centering
\caption{Effects of the finetuning data to the performance in downstream tasks. We use 4-bit weight 6-bit activation LLaMA-7B for the experiments. We test three strategies for data generation. Generated data$^1$ refers to always picking the top-1 candidate without sampling. Generated data$^2$ refers to sampling the next token from the distribution. Generated data$^3$ refers to first 3\textasciitilde5 tokens are generated with deterministic selection while the rest are stochastically sampled from the distribution.}
\label{tab:ablation_data}
\setlength{\tabcolsep}{1.2mm}
{\resizebox{0.98\textwidth}{!}{
\begin{tabular}{ll:cc:ccccccccc}
\noalign{\vspace{0.4em}}\hline\noalign{\vspace{0.1em}}
 & &  \textbf{C4} & \textbf{Wiki2} & \textbf{BoolQ} & \textbf{PIQA} & \textbf{SIQA} & \textbf{HellaSwag} & \textbf{WinoGrande} & \textbf{ARC-e} & \textbf{ARC-c} & \textbf{OBQA} & \textbf{Avg.}\\ 
& \textbf{Finetuning Data} & ($\downarrow$) & ($\downarrow$) & ($\uparrow$) & ($\uparrow$)& ($\uparrow$)& ($\uparrow$)& ($\uparrow$)& ($\uparrow$)& ($\uparrow$)& ($\uparrow$)& ($\uparrow$) \\
\noalign{\vspace{0.1em}}\hline\noalign{\vspace{0.1em}}
\tiny{1} & (Pretrained Model) & 7.2 & 10.7 & 76.8 & 79.3 & 48.6 & 76.1 & 70.0 & 73.0 & 48.0 & 57.6 & 66.2 \\
\noalign{\vspace{0.1em}}\hdashline[0.8pt/1pt]\noalign{\vspace{0.1em}}
\tiny{2} & wiki2              & 10.1 & 5.5 & 46.9 & 74.3 & 45.2 & 72.4 & 65.7 & 67.2 & 45.0 & 47.8 & 58.1  \\
\tiny{3} & wiki103            & 9.6 & \textbf{5.2 }& 45.9 & 74.4 & 46.4 & 71.4 & 66.1 & 67.5 & 46.3 & 49.8 & 58.5  \\
\tiny{4} & c4                 & 7.8 & 11.3 & 61.7 & 77.7 & 48.8 & 73.2 & 67.2 & 67.8 & 43.6 & 52.2 & 61.5  \\
\tiny{5} & Generated data$^1$ & 8.0 & 11.4 & 60.0 & 77.1 & 48.1 & 72.3 & 65.7 & 67.4 & 44.2 & 49.8 & 60.6  \\
\tiny{6} & Generated data$^2$ & 7.7 & 11.5 & 70.9 & 76.1 & 47.9 & 72.2 & 66.9 & 69.3 & 46.4 & 53.6 & 62.9  \\
\tiny{7} & Generated data$^3$ & \textbf{7.7} & 10.8 & 72.9 & 76.8 & 47.9 & 72.4 & 68.3 & 68.8 & 44.2 & 53.2 & \textbf{63.1} \\
\hline
\end{tabular}}}
\end{table}

\subsubsection{Quantization Function}
\label{sec:ablation_quant}
We compare the no-clipping quantization method with clipping-based methods in Table~\ref{tab:ablation_quant}. Following the practice in previous works~\citep{liu2022bit,liu2023oscillation}, we use StatsQ~\citep{liu2022nonuniform}, a statistically-calculated scaling factor for clipping-based weight quantization and LSQ~\citep{esser2019learned}, the learnable scaling factor for clipping-based activation quantization. However, our findings indicate that these two state-of-the-art clipping-based quantization methods do not surpass the performance achieved by the MinMax non-clipping method. This observation reinforces the argument that preserving the outliers is critical to the performance of large language models.

Furthermore, we observe that for LLaMA models, the activations and weights exhibit predominantly symmetric distributions, which makes using symmetric quantizers the best choice. It is important to note, however, that this conclusion may not hold true for other large language models, especially those incorporating GeLU layers.
\begin{table}[t]
\renewcommand\arraystretch{0.6}
\centering
\caption{Ablation study on the effects of the quantization methods on LLaMA-7B model. The quantization level is set to 4-bit weight and 8-bit activation. } 
\label{tab:ablation_quant}
\setlength{\tabcolsep}{1.2mm}
{\resizebox{0.98\textwidth}{!}{
\begin{tabular}{lll:cc:cccccccccc}
\vspace{-1em}
& & & & & & & &\\
\noalign{\vspace{0.5em}}\hline\noalign{\vspace{0.1em}}
& & & \textbf{C4} & \textbf{Wiki2} & \textbf{BoolQ} & \textbf{PIQA} & \textbf{SIQA} & \textbf{HellaSwag} & \textbf{WinoGrande} & \textbf{ARC-e} & \textbf{ARC-c} & \textbf{OBQA} & \textbf{Avg.}\\ 
& \textbf{Weight} & \textbf{Activation}  & ($\downarrow$) & ($\downarrow$) & ($\uparrow$) & ($\uparrow$)& ($\uparrow$)& ($\uparrow$)& ($\uparrow$)& ($\uparrow$)& ($\uparrow$)& ($\uparrow$)& ($\uparrow$) \\
\noalign{\vspace{0.1em}}\hline\noalign{\vspace{0.1em}}
\tiny{1} & \multicolumn{2}{l}{(Pretrained Model)} & 7.2 & 10.7 & 76.8 & 79.3 & 48.6 & 76.1 & 70.0 & 73.0 & 48.0 & 57.6 & 66.2 \\
           \noalign{\vspace{0.1em}}\hdashline[0.8pt/1pt]\noalign{\vspace{0.1em}}
\tiny{2} & StatsQ & LSQ & 9.0 & 11.9 & 64.9 & 66.8 & 43.6 & 63.5 & 56.1 & 51.0 & 31.4 & 33.8 & 51.4 \\
\tiny{3} & MinMax & LSQ  & 9.4 & 12.8 & 63.5 & 62.4 & 42.4 & 61.2 & 52.9 & 45.6 & 29.6 & 33.8 & 48.9 \\
\tiny{4} & StatsQ & MinMax & 8.2 & 11.0 & 71.7 & 75.1 & 43.7 & 69.5 & 58.9 & 62.6 & 35.2 & 37.8 & 56.8 \\
\tiny{5} & MinMax & MinMax  & \textbf{7.4} & \textbf{10.9} & 74.8 & 77.8 & 48.6 & 73.6 & 69.0 & 69.7 & 45.8 & 55.8 & \textbf{64.4} \\
           \noalign{\vspace{0.1em}}\hdashline[0.8pt/1pt]\noalign{\vspace{0.1em}}
\tiny{6} & Asym &  Asym  & 7.3 & 10.4 & 75.0 & 78.4 & 48.0 & 73.9 & 69.3 & 71.9 & 45.7 & 52.6 & 64.3 \\
\tiny{7} & Sym &  Asym  & 7.4 & 11.0 & 72.7 & 77.9 & 48.8 & 73.3 & 67.9 & 69.2 & 45.2 & 56.0 & 63.9 \\
\tiny{8} & Asym &  Sym &  7.4 & 10.9 & 73.3 & 78.4 & 48.0 & 73.9 & 68.9 & 71.4 & 46.4 & 54.0 & 64.3 \\
\tiny{9} & Sym & Sym & \textbf{7.4} & \textbf{10.9} & 74.8 & 77.8 & 48.6 & 73.6 & 69.0 & 69.7 & 45.8 & 55.8 & \textbf{64.4} \\
\hline
\end{tabular}}}
\end{table}

\subsubsection{Knowledge Distillation}
\label{sec:ablation_kd}
Table~\ref{tab:ablation_kd} shows that different knowledge distillation methods have a significant impact on the final accuracy of fine-tuned models. Notably, utilizing the next token alone as the label is sub-optimal due to the inherent randomness and noise introduced by sampling from a distribution of candidates during the generation process.
In contrast, logit distillation, which utilizes the complete logit distribution prediction from the teacher model, leads to superior performance of fine-tuned models compared to label-based training approaches. Interestingly, we have observed that incorporating attention distillation or hidden layer distillation actually hampers the performance. Consequently, we exclusively employ logit distillation in all our experiments.

\begin{table}[t]
\renewcommand\arraystretch{0.6}
\centering
\caption{Ablation study on the knowledge distillation choices on LLaMA-7B model with generated data. The quantization level is set to 4-bit weight and 6-bit activation.}
\label{tab:ablation_kd}
\setlength{\tabcolsep}{1.2mm}
{\resizebox{0.98\textwidth}{!}{
\begin{tabular}{ll:cc:cccccccccc}
\vspace{-0.6em}
& & & & & & & &\\
\hline\noalign{\vspace{0.1em}}
& & \textbf{C4} & \textbf{Wiki2} & \textbf{BoolQ}  & \textbf{PIQA} & \textbf{SIQA} & \textbf{HellaSwag} & \textbf{WinoGrande} & \textbf{ARC-e} & \textbf{ARC-c} & \textbf{OBQA} & \textbf{Avg.}\\
& \textbf{Method} & ($\downarrow$) & ($\downarrow$) & ($\uparrow$) & ($\uparrow$)& ($\uparrow$)& ($\uparrow$)& ($\uparrow$)& ($\uparrow$)& ($\uparrow$)& ($\uparrow$)& ($\uparrow$) \\
\noalign{\vspace{0.1em}}\hline\noalign{\vspace{0.1em}}
\noalign{\vspace{0.1em}}
\tiny{1} & (Pretrained Model) & 7.2 & 10.4 & 76.8 & 79.3 & 48.6 & 76.1 & 70.0 & 73.0 & 48.0 & 57.6 & 66.2 \\
\noalign{\vspace{0.1em}}\hdashline[0.8pt/1pt]\noalign{\vspace{0.1em}}
\tiny{2} & Label  & 8.1 & 11.9 & 69.4 & 77.3 & 48.7 & 72.1 & 67.1 & 67.6 & 45.4 & 51.4 & 62.4 \\
\tiny{3} & Lable + Attention  & 8.8 & 18.6 & 70.2 & 75.3 & 47.6 & 68.9 & 67.2 & 65.6 & 42.6 & 51.2 & 61.1 \\
\tiny{4} & Lable + Hidden  & 10.9 & 16.2 & 61.0 & 53.5 & 41.1 & 32.6 & 50.2 & 25.8 & 23.1 & 25.0 & 37.7 \\
\tiny{5} & Lable + Logits  & 7.8 & 11.0 & 70.8 & 77.3 & 48.3 & 72.5 & 66.7 & 68.2 & 46.5 & 55.4 & \textbf{63.2} \\
\tiny{6} & \textbf{Logits}  & \textbf{7.7}& \textbf{10.8}& 72.9 & 76.8 & 47.9 & 72.4 & 68.3 & 68.8 & 44.2 & 53.2 & \textbf{63.1} \\
\tiny{7} & Logits + Attention  & 7.9 & 12.2 & 73.2 & 74.6 & 47.2 & 69.1 & 65.1 & 64.8 & 42.1 & 52.8 & 61.1 \\
\tiny{8} & Logits + Hidden  & 22.3 & 52.6 & 38.0 & 50.4 & 38.6 & 25.6 & 50.5 & 26.3 & 24.3 & 25.8 & 34.9 \\
\tiny{9} & Logits + Hidden + Attention  & 21.9 & 46.0 & 55.0 & 47.8 & 39.0 & 33.4 & 48.5 & 29.7 & 26.4 & 25.8 & 38.2 \\
\hline
\end{tabular}}}
\end{table}
\subsection{Compatibility with SmoothQuant}\label{sec:smoothquant}
Our method is also compatible with the weight activation rescale technique proposed in SmoothQuant~\citep{xiao2022smoothquant}.
Table~\ref{tab:smoothquant} shows that incorporating SmoothQuant into 4-bit weight 4-bit activation (W4A4) quantization can further improve accuracy. However, in the case where the activation bit is greater than the weight bit (\textit{i.e,} W4A8), adding SmoothQuant does not yield any improvement and may even harm the performance.

\begin{table}[t]
\renewcommand\arraystretch{0.6}
\centering
\caption{Combine $\ours{}$ with SmoothQuant for lower quantization bit settings}
\label{tab:smoothquant}
\setlength{\tabcolsep}{1.2mm}
{\resizebox{0.98\textwidth}{!}{
\begin{tabular}{llcccccccccccc}
\noalign{\vspace{0.3em}}
\hline
\noalign{\vspace{0.1em}}
& & & \textbf{BoolQ} & \textbf{PIQA} & \textbf{SIQA} & \textbf{HellaSwag} & \textbf{WinoGrande} & \textbf{ARC-e} & \textbf{ARC-c} & \textbf{OBQA} & \textbf{Avg.} \\ 
& \textbf{Method} & \textbf{\#Bits} & ($\uparrow$) & ($\uparrow$)& ($\uparrow$)& ($\uparrow$)& ($\uparrow$)& ($\uparrow$)& ($\uparrow$)& ($\uparrow$)& ($\uparrow$) \\
\hline
\noalign{\vspace{0.1em}}
\tiny{1} & LLaMA-7B & 16-16-16 & 76.8 & 79.3 & 48.6 & 76.1 & 70.0 & 73.0 & 48.0 & 57.6 & 66.2 \\
           \noalign{\vspace{0.1em}}\hdashline[0.8pt/1pt]\noalign{\vspace{0.1em}}
\tiny{2} & RTN & 4-8-16 & 67.6 & 77.4 & 47.1 & 71.6 & 66.9 & 67.1 & 45.8 & 52.0 & 61.9 \\
\tiny{3} & SmoothQuant & 4-8-16 & 70.2 & 76.4 & 44.8 & 68.1 & 66.0 & 67.3 & 42.9 & 49.0 & 60.6 \\
\tiny{4} & \ours{} & 4-8-16 & 74.8 & 77.8 & 48.6 & 73.6 & 69.0 & 69.7 & 45.8 & 55.8 & \textbf{64.4} \\
\tiny{5} & \ours{} + SmoothQuant & 4-8-16 & 74.1 & 77.2 & 47.8 & 71.9 & 67.7 & 69.6 & 44.8 & 55.4 & 63.6 \\
           \noalign{\vspace{0.1em}}\hdashline[0.8pt/1pt]\noalign{\vspace{0.1em}} 
\tiny{6} & RTN & 4-4-16 & 51.3 & 49.8 & 36.9 & 26.2 & 47.9 & 25.7 & 24.5 & 31.2 & 36.7 \\
\tiny{7} & SmoothQuant & 4-4-16 & 54.1 & 62.8 & 41.8 & 41.5 & 52.6 & 50.6 & 32.9 & 36.4 & 46.6 \\
\tiny{8} & \ours{} & 4-4-16 & 57.9 & 47.5 & 39.9 & 25.8 & 47.6 & 27.2 & 25.8 & 29.4 & 37.6 \\
\tiny{9} & \ours{} + SmoothQuant & 4-4-16 & 63.5 & 64.3 & 41.8 & 55.6 & 52.9 & 50.3 & 30.2 & 35.0 & \textbf{49.2} \\
           \noalign{\vspace{0.1em}}\hdashline[0.8pt/1pt]\noalign{\vspace{0.1em}} 
\tiny{10} & RTN & 4-4-4 & 50.2 & 50.5 & 37.1 & 26.0 & 49.6 & 26.1 & 24.4 & 28.6 & 36.6 \\
\tiny{11} & SmoothQuant & 4-4-4 & 49.1 & 49.8 & 39.1 & 27.4 & 48.0 & 30.4 & 25.8 & 29.2 & 37.4 \\
\tiny{12} & \ours{} & 4-4-4 & 61.3 & 51.5 & 39.2 & 31.1 & 51.9 & 27.9 & 23.9 & 29.4 & 39.5 \\
\tiny{13} & \ours{} + SmoothQuant & 4-4-4 & 62.4 & 55.9 & 40.9 & 47.8 & 50.6 & 35.5 & 26.4 & 34.6 & \textbf{44.3} \\
\hline
\end{tabular}}}
\end{table}

\section{Related Works}

\textbf{Quantization}
Neural network quantization is proved to be a valuable tool in compressing model size and reducing storage consumption. Classic quantization methods, such as MinMax quantization~\citep{jacob2018quantization,krishnamoorthi2018quantizing}, Learned step-size quantization~\citep{esser2019learned}, PACT~\citep{choi2018pact}, N2UQ~\citep{liu2022nonuniform} and etc, have primarily been developed for convolutional neural networks. While several recent works have explored language model compression, they are mostly focused on smaller models~\citep{zafrir2019q8bert, fan2020training,shen2020q,zadeh2020gobo,bai2021binarybert,qin2021bibert,liu2022bit} like BERT~\citep{devlin2019bert} or BART~\citep{lewis2019bart}. For large language models (LLMs), the available quantization methods are mostly limited to post-training quantization~\citep{xiao2022smoothquant,yao2022zeroquant,frantar2022gptq}, due to the lack of accessible training data or the prohibitive resource requirements for fine-tuning on the entire pre-training dataset. To the best of our knowledge, no previous work has addressed the specific challenge of quantization-aware training for LLMs.

\textbf{Data generation}
Data generation for QAT remains a relatively unexplored field of research. While there are several works in the vision domain fine-tuning student networks~\citep{yin2020dreaming,liu2022data,cai2020zeroq} using data generated by pre-trained teacher models, these methods mainly focus on image data. They involve updating the noise input based on gradients computed from the label and reconstructing images by accumulating these gradients. In contrast, our proposed approach introduces next token data generation in the language domain. This method is more natural and proves to be effective for fine-tuning quantized language models.

\section{Conclusion and Limitations}
We proposed data-free quantization-aware training for LLMs and showed accurate, 4-bit quantization is possible using this technique.  Given the generality of the training-data-agnostic distillation method, and the growing cost of LLM deployments, we expect our method to have wide applicability.  For instance, the method could also be used for models trained in several stages, e.g. with instruction tuning or reinforcement learning~\citep{instruct-gpt}.  We leave this investigation to future work.
Since 4-bit quantization does not have hardware support out-of-the-box, we haven't included hardware implementation as part of this work.  However, we're working with partners to enable this in the near future.
While our method works well for 4-bit weights, 4-bit KV cache and 8-bit activations, it isn't sufficient for 4-bit activation quantization.  This case needs further research.

\bibliography{main}
\bibliographystyle{references}

\newpage
\appendix
\section{Appendix}

\subsection{Few-shot Evaluation Results}
Table~\ref{tab:few-shot} presents the few-shot performance of the quantized model on the MMLU~\cite{hendrycks2020mmlu} and TriviaQA~\cite{joshi2017triviaqa} benchmarks.

\begin{table}[h]
\renewcommand\arraystretch{0.6}
\centering
\caption{5-shot few-shot exact match performance on the TriviaQA dataset~\cite{} and 5-shot accuracy on Massive Multitask Language Understanding (MMLU) dataset~\cite{}. }
\label{tab:few-shot}
\setlength{\tabcolsep}{0.8mm}
{\resizebox{0.85\textwidth}{!}{
\begin{tabular}{llcc:ccccc:c}
\vspace{-4.5em}
& & & & & & & &\\
& & & & & & & &\\
& & & & & & & &\\
& & & & & & & &\\
& & & & & & & &\\
& & & & & & & &\\
& & & & & & & &\\
\hline
& & & & \multicolumn{5}{c}{\textbf{MMLU}} & \\
& & & & \textbf{Humanities} & \textbf{STEM} & \textbf{Social Sciences} & \textbf{Other} & \textbf{Average} & \textbf{TriviaQA} \\ 
& \textbf{Method} & \textbf{\#Bits} &  \begin{tabular}[c]{@{}c@{}}\textbf{Size}$_\text{ (GB)}$\end{tabular}  & ($\uparrow$) & ($\uparrow$) & ($\uparrow$) & ($\uparrow$) & ($\uparrow$) & ($\uparrow$)\\
\hline
\noalign{\vspace{0.1em}}
\tiny{1}& LLaMA-7B & 16-16-16 & 12.6 & 33.5 & 30.6 & 38.4 & 39.1 & 35.2 & 57.0 \\
            \noalign{\vspace{0.1em}}\hdashline[0.8pt/1pt]\noalign{\vspace{0.1em}} 
\tiny{2}  & RTN & 4-8-4 & 3.5 & 23.9 & 26.8 & 26.5 & 24.4 & 25.2 & 0.3 \\
\tiny{3}  & SmoothQuant & 4-8-4 & 3.5 & 24.3 & 27.5 & 26.2 & 24.6 & 25.5 & 3.9 \\
\tiny{4}  & \ours{} & 4-8-4 & 3.5 & 25.6 & 24.3 & 24.0 & 27.8 & \textbf{25.5} & \textbf{42.6} \\
            \noalign{\vspace{0.1em}}\hdashline[0.8pt/1pt]\noalign{\vspace{0.1em}} 
\tiny{5}  & RTN & 4-8-8 & 3.5 & 30.1 & 25.6 & 27.5 & 32.5 & 29.1 & 44.5 \\
\tiny{6}  & SmoothQuant & 4-8-8 & 3.5 & 27.1 & 28.9 & 28.0 & 31.9 & 28.7 & 39.6 \\
\tiny{7}  & \ours{} & 4-8-8 & 3.5 & 30.0 & 27.4 & 28.4 & 34.2 & \textbf{30.0} & \textbf{50.8} \\
            \noalign{\vspace{0.1em}}\hdashline[0.8pt/1pt]\noalign{\vspace{0.1em}} 
\tiny{8 } & RTN & 4-6-16 & 3.5 & 27.0 & 26.0 & 25.8 & 27.0 & 26.5 & 36.0 \\
\tiny{9 } & SmoothQuant & 4-6-16 & 3.5 & 26.2 & 27.0 & 27.5 & 29.9 & 27.5 & 36.2 \\
\tiny{10} & \ours{} & 4-6-16 & 3.5 & 28.9 & 27.3 & 31.6 & 33.0 & \textbf{30.0} & \textbf{49.0} \\
            \noalign{\vspace{0.1em}}\hdashline[0.8pt/1pt]\noalign{\vspace{0.1em}} 
\tiny{11} & RTN & 4-8-16 & 3.5 & 30.2 & 25.9 & 26.8 & 32.0 & 28.9 & 44.9 \\
\tiny{12} & SmoothQuant & 4-8-16 & 3.5 & 26.9 & 28.6 & 29.6 & 32.0 & 29.0 & 40.0 \\
\tiny{13} & \ours{} & 4-8-16 & 3.5 & 30.3 & 28.1 & 30.3 & 34.5 & \textbf{30.8} & \textbf{50.8} \\
            \noalign{\vspace{0.1em}}\hdashline[0.8pt/1pt]\noalign{\vspace{0.1em}} 
\tiny{14} & RTN & 8-8-4 & 6.5 & 24.2 & 27.3 & 25.8 & 24.5 & 25.3 & 14.8 \\
\tiny{15} & SmoothQuant & 8-8-4 & 6.5 & 24.4 & 26.4 & 25.6 & 24.2 & 25.1 & 32.8 \\
\tiny{16} & \ours{} & 8-8-4 & 6.5 & 28.3 & 25.5 & 28.7 & 30.4 & \textbf{28.2} & \textbf{46.2} \\
            \noalign{\vspace{0.1em}}\hdashline[0.8pt/1pt]\noalign{\vspace{0.1em}} 
\tiny{17} & RTN & 8-8-8 & 6.5 & 34.3 & 31.9 & 38.5 & 40.5 & \textbf{36.1} & 56.6 \\
\tiny{18} & SmoothQuant & 8-8-8 & 6.5 & 33.2 & 31.5 & 38.5 & 38.9 & 35.3 & \textbf{56.7} \\
\tiny{19} & \ours{} & 8-8-8 & 6.5 & 32.9 & 29.7 & 37.9 & 37.9 & 34.4 & 56.1 \\
            \noalign{\vspace{0.1em}}\hdashline[0.8pt/1pt]\noalign{\vspace{0.1em}} 
\tiny{20} & RTN & 8-8-16 & 6.5 & 34.4 & 31.8 & 39.3 & 39.9 & \textbf{36.1} & 56.6 \\
\tiny{21} & SmoothQuant & 8-8-16 & 6.5 & 33.0 & 30.5 & 38.7 & 38.8 & 35.0 & \textbf{56.8} \\
\tiny{22} & \ours{} & 8-8-16 & 6.5 & 32.2 & 29.4 & 37.0 & 37.6 & 33.8 & 56.1 \\
            \noalign{\vspace{0.1em}}\hline\noalign{\vspace{0.1em}} 
\tiny{23} & LLaMA-13B & 16-16-16 & 24.2 & 44.4 & 36.2 & 54.3 & 53.3 & 46.7 & 63.7 \\
            \noalign{\vspace{0.1em}}\hdashline[0.8pt/1pt]\noalign{\vspace{0.1em}} 
\tiny{24} & RTN & 4-8-4 & 6.5 & 25.5 & 24.9 & 24.3 & 26.5 & 25.3 & 22.2 \\
\tiny{25} & SmoothQuant & 4-8-4 & 6.5 & 25.6 & 22.8 & 23.4 & 26.4 & 24.7 & 32.7 \\
\tiny{26} & \ours{} & 4-8-4 & 6.5 & 29.4 & 28.5 & 31.9 & 35.8 & \textbf{31.1} & \textbf{54.3} \\
            \noalign{\vspace{0.1em}}\hdashline[0.8pt/1pt]\noalign{\vspace{0.1em}} 
\tiny{27} & RTN & 4-8-8 & 6.5 & 38.3 & 32.7 & 45.3 & 46.4 & 40.4 & 57.9 \\
\tiny{28} & SmoothQuant & 4-8-8 & 6.5 & 30.9 & 28.6 & 33.4 & 37.1 & 32.3 & 46.6 \\
\tiny{29} & \ours{} & 4-8-8 & 6.5 & 38.7 & 32.8 & 47.1 & 47.7 & \textbf{41.2} & \textbf{59.3} \\
            \noalign{\vspace{0.1em}}\hdashline[0.8pt/1pt]\noalign{\vspace{0.1em}} 
\tiny{30} & RTN & 4-6-16 & 6.5 & 28.5 & 27.8 & 29.5 & 32.0 & 29.3 & 39.6 \\
\tiny{31} & SmoothQuant & 4-6-16 & 6.5 & 30.3 & 29.6 & 33.5 & 37.1 & 32.4 & 44.8 \\
\tiny{32} & \ours{} & 4-6-16 & 6.5 & 37.4 & 33.4 & 45.1 & 46.0 & \textbf{40.1} & \textbf{57.7} \\
            \noalign{\vspace{0.1em}}\hdashline[0.8pt/1pt]\noalign{\vspace{0.1em}} 
\tiny{33} & RTN & 4-8-16 & 6.5 & 38.7 & 32.6 & 45.2 & 45.8 & 40.3 & 57.9 \\
\tiny{34} & SmoothQuant & 4-8-16 & 6.5 & 30.3 & 27.8 & 34.3 & 37.5 & 32.2 & 46.6 \\
\tiny{35} & \ours{} & 4-8-16 & 6.5 & 40.1 & 32.4 & 47.6 & 48.0 & \textbf{41.8} & \textbf{59.8} \\
            \noalign{\vspace{0.1em}}\hdashline[0.8pt/1pt]\noalign{\vspace{0.1em}} 
\tiny{36} & RTN & 8-8-4 & 12.4 & 27.8 & 26.2 & 27.0 & 29.6 & 27.6 & 44.3 \\
\tiny{37} & SmoothQuant & 8-8-4 & 12.4 & 27.8 & 28.1 & 28.6 & 32.3 & 29.1 & 49.6 \\
\tiny{38} & \ours{} & 8-8-4 & 12.4 & 34.1 & 29.3 & 38.7 & 40.7 & \textbf{35.5} & \textbf{58.8} \\
            \noalign{\vspace{0.1em}}\hdashline[0.8pt/1pt]\noalign{\vspace{0.1em}} 
\tiny{39} & RTN & 8-8-8 & 12.4 & 44.2 & 35.6 & 52.2 & 52.5 & 45.9 & 62.9 \\
\tiny{40} & SmoothQuant & 8-8-8 & 12.4 & 44.5 & 36.1 & 53.5 & 53.3 & \textbf{46.6} & \textbf{63.4} \\
\tiny{41} & \ours{} & 8-8-8 & 12.4 & 43.5 & 36.1 & 52.6 & 52.5 & 45.8 & 63.3 \\
            \noalign{\vspace{0.1em}}\hdashline[0.8pt/1pt]\noalign{\vspace{0.1em}} 
\tiny{42} & RTN & 8-8-16 & 12.4 & 44.3 & 34.9 & 51.7 & 53.0 & 45.7 & 63.1 \\
\tiny{43} & SmoothQuant & 8-8-16 & 12.4 & 44.5 & 36.4 & 53.7 & 53.4 & \textbf{46.7} & \textbf{63.4} \\
\tiny{44} & \ours{} & 8-8-16 & 12.4 & 43.6 & 36.1 & 53.8 & 53.2 & 46.3 & \textbf{63.4} \\
            \noalign{\vspace{0.1em}}\hline\noalign{\vspace{0.1em}} 
\tiny{23} & LLaMA-30B & 16-16-16 & 60.6 & 55.8 & 46.0 & 66.7 & 63.4 & 57.8 & 69.9 \\
            \noalign{\vspace{0.1em}}\hdashline[0.8pt/1pt]\noalign{\vspace{0.1em}} 
\tiny{46} & RTN & 4-8-4 & 15.7 & 24.4 & 26.2 & 27.2 & 26.4 & 25.9 & 19.2\\
\tiny{47} & SmoothQuant & 4-8-4 & 15.7 & 23.9 & 27.5 & 23.2 & 24.1 & 24.6 & 7.5\\
\tiny{48} & \ours{} & 4-8-4 & 15.7 & 47.6 & 40.4 & 55.9 & 54.5 & \textbf{49.3} & \textbf{63.5}\\
          \noalign{\vspace{0.1em}}\hdashline[0.8pt/1pt]\noalign{\vspace{0.1em}} 
\tiny{49} & RTN & 4-8-8 & 15.7 & 51.0 & 43.6 & 62.2 & 60.6 & 53.9 & \textbf{66.8}\\
\tiny{50} & SmoothQuant & 4-8-8 & 15.7 & 35.2 & 35.1 & 46.9 & 45.2 & 40.0 & 57.9\\
\tiny{51} & \ours{} & 4-8-8 & 15.7 & 52.2 & 44.3 & 61.4 & 61.0 & \textbf{54.4} & 65.9\\
          \noalign{\vspace{0.1em}}\hdashline[0.8pt/1pt]\noalign{\vspace{0.1em}} 
\tiny{52} & RTN & 4-6-16 & 15.7 & 29.5 & 31.3 & 32.1 & 36.2 & 32.0 & 39.3\\
\tiny{53} & SmoothQuant & 4-6-16 & 15.7 & 31.6 & 34.3 & 43.4 & 42.3 & 37.2 & 56.7\\
\tiny{54} & \ours{} & 4-6-16 & 15.7 & 47.7 & 41.7 & 58.9 & 57.5 & \textbf{51.0} & \textbf{64.2}\\
          \noalign{\vspace{0.1em}}\hdashline[0.8pt/1pt]\noalign{\vspace{0.1em}} 
\tiny{55} & RTN & 4-8-16 & 15.7 & 50.9 & 44.0 & 62.8 & 61.3 & 54.2 & \textbf{67.1}\\
\tiny{56} & SmoothQuant & 4-8-16 & 15.7 & 35.6 & 36.2 & 48.6 & 45.7 & 40.8 & 58.5\\
\tiny{57} & \ours{} & 4-8-16 & 15.7 & 52.8 & 44.4 & 63.6 & 61.2 & \textbf{55.1} & \textbf{67.1}\\
          \noalign{\vspace{0.1em}}\hdashline[0.8pt/1pt]\noalign{\vspace{0.1em}} 
\tiny{58} & RTN & 8-8-4 & 30.7 & 26.1 & 27.6 & 28.6 & 29.0 & 27.6 & 30.2\\
\tiny{59} & SmoothQuant & 8-8-4  & 30.7& 27.9 & 29.1 & 31.7 & 33.1 & 30.1 & 38.9\\
\tiny{60} & \ours{} & 8-8-4 & 30.7 & 49.7 & 42.2 & 60.8 & 59.7 & \textbf{52.7} & \textbf{67.9}\\
          \noalign{\vspace{0.1em}}\hdashline[0.8pt/1pt]\noalign{\vspace{0.1em}} 
\tiny{61} & RTN & 8-8-8 & 30.7 & 55.6 & 45.8 & 66.3 & 63.4 & 57.5 & \textbf{70.4}\\
\tiny{62} & SmoothQuant & 8-8-8 & 30.7 & 56.0 & 46.0 & 67.3 & 64.1 & 58.0 & 70.2\\
\tiny{63} & \ours{} & 8-8-8 & 30.7 & 56.5 & 47.7 & 66.9 & 64.2 & \textbf{58.5} & 69.4\\
          \noalign{\vspace{0.1em}}\hdashline[0.8pt/1pt]\noalign{\vspace{0.1em}} 
\tiny{64} & RTN & 8-8-16 & 30.7 & 56.3 & 45.6 & 66.8 & 63.7 & 57.8 & \textbf{70.3}\\
\tiny{65} & SmoothQuant & 8-8-16 & 30.7 & 56.0 & 46.7 & 67.5 & 63.8 & \textbf{58.2} & \textbf{70.3}\\
\tiny{66} & \ours{} & 8-8-16 & 30.7 & 54.9 & 45.9 & 66.7 & 63.6 & 57.4 & 70.0\\
\hline
\end{tabular}}}
\vspace{-2em}
\end{table}

\subsection{Memory consumption of KV cache}
We compute the memory required to store the key-value cache (KV cache) in large language models for various sequence lengths, as shown in Table~\ref{tab:kvcache}. It is evident that the size of the KV cache can quickly exceed the model size when dealing with longer sequences. Given the increasing use of long inputs and contexts in various applications, it becomes crucial to compress the KV cache.
\begin{table}[t]
\renewcommand\arraystretch{0.6}
\centering
\caption{The memory consumption of key-value cache (KV cache) for different sequence length.}
\label{tab:kvcache}
\setlength{\tabcolsep}{1.2mm}
{\resizebox{0.8\textwidth}{!}{
\begin{tabular}{llcccccccccccc}
\noalign{\vspace{0.3em}}
\hline
\noalign{\vspace{0.1em}}
& \multicolumn{3}{c}{\textbf{LLaMA-7B}} & \multicolumn{3}{c}{\textbf{LLaMA-13B}} & \multicolumn{3}{c}{\textbf{LLaMA-30B}} \\ 
\textbf{Text Length} & 16-bit & 8-bit & 4-bit & 16-bit & 8-bit & 4-bit & 16-bit & 8-bit & 4-bit \\
\noalign{\vspace{0.1em}}\hline\noalign{\vspace{0.1em}}
1k  & 0.25 GB & 0.13 GB & 0.06 GB & 0.39 GB & 0.20 GB & 0.10 GB & 0.76 GB & 0.38 GB & 0.19 GB \\
2k  & 0.50 GB & 0.25 GB & 0.13 GB & 0.78 GB & 0.39 GB & 0.20 GB & 1.52 GB & 0.76 GB & 0.38 GB \\
4k  & 1.00 GB & 0.50 GB & 0.25 GB & 1.56 GB & 0.78 GB & 0.39 GB & 3.05 GB & 1.52 GB & 0.76 GB \\
8k  & 2.00 GB & 1.00 GB & 0.50 GB & 3.13 GB & 1.56 GB & 0.78 GB & 6.09 GB & 3.05 GB & 1.52 GB \\
16k & 4.00 GB & 2.00 GB & 1.00 GB & 6.25 GB & 3.13 GB & 1.56 GB & 12.19 GB & 6.09 GB & 3.05 GB \\
32k & 8.00 GB & 4.00 GB & 2.00 GB & 12.50 GB & 6.25 GB & 3.13 GB & 24.38 GB & 12.19 GB & 6.09 GB \\
\hline
\end{tabular}}}
\end{table}

\subsection{Evaluation Benchmarks}

\subsubsection{Zero-shot Common Sense Reasoning tasks}
\textbf{BoolQ}~\citep{clark2019boolq} is a reading comprehension dataset of naturally occurring yes/no questions. Each example consists of a question (Q), an excerpt from a passage (P), and an answer (A) with an explanation added for clarity.

\textbf{PIQA}~\citep{bisk2020piqa}, short for Physical
Interaction: Question Answering, is a benchmark for evaluating
and studying physical commonsense understanding in natural language models. 

\textbf{SIQA}~\citep{sap2019siqa} aims to measure the social and emotional intelligence of computational models through multiple choice question answering (QA).

\textbf{HellaSwag}~\citep{zellers2019hellaswag} is a benchmark for physically situated commonsense natural language inference. It consists the four-way multiple-choice problems that are trivial for humans (> 95\% accuracy), but challenging for the language models.

\textbf{WinoGrande}~\citep{sakaguchi2021winogrande} is a benchmark for commonsense reasoning. It comprises a set of 273 expert-crafted pronoun resolution problems originally designed to be unsolvable for statistical models that rely on selectional preferences or word associations.

\textbf{ARC}~\citep{clark2018arc}, the AI2 Reasoning Challenge, contains a collection of 7787 natural science questions. It is partitioned into a Challenge Set and an Easy Set, where the Challenge Set contains only questions answered incorrectly by both a retrieval-based algorithm and a word co-occurrence algorithm. 

\textbf{OBQA}~\citep{mihaylov2018obqa} is a dataset of about 6000 questions for open book question answering. The task focuses on the challenge of combining a corpus of provided science facts (open book) with external broad common knowledge. 

\subsubsection{Few-shot Tasks}
\textbf{TriviaQA}~\citep{joshi2017triviaqa} is a closed-book question answering
benchmark. It contains over 650K question-answer evidence triples, that are derived by combining 95K Trivia enthusiast authored question-answer pairs with on average six supporting evidence documents per question.

\textbf{MMLU~}\citep{hendrycks2020mmlu}, the Massive Multitask Language Understanding(MMLU) benchmark~\cite{hendrycks2020mmlu}, consists of multiple choice questions covering various domains of knowledge, including humanities, STEM and social sciences. 

\subsection{Generation Tasks}
\textbf{WikiText2}~\citep{merity2016wiki2} is a collection of over 100 million tokens extracted from the set of verified Good and Featured articles on Wikipedia. 

\textbf{C4}~\citep{raffel2020c4}, abbreviate for Colossal Clean Crawled Corpus. Common Crawl\footnote{http://commoncrawl.org/} is a publicly-available web archive that provides “web extracted text” by removing markup and other non-text content from the scraped HTML files. C4 filters Common Crawl’s web-extracted text and produces a collection of text that comprises clean and natural English text.

\end{document}